# DeepJIVE: Learning Joint and Individual Variation Explained from Multimodal Data Using Deep Learning

Matthew Drexler[1], Benjamin Risk[1], James J Lah[1,2], Suprateek Kundu[3], Deqiang Qiu[1], for the Alzheimer's Disease Neuroimaging Initiative[*]

[1]Emory University School of Medicine, [2]Emory University Hospital, [3]The University of Texas MD Anderson Cancer Center

**Abstract**

Conventional multimodal data integration methods provide a comprehensive assessment of the shared and/or unique structure within each individual data type but suffer from several limitations such as the inability to handle high-dimensional data and identify nonlinear structures. In this paper, we introduce DeepJIVE, a deep-learning approach to performing Joint and Individual Variance Explained (JIVE). We perform mathematical derivation and experimental validations using both synthetic and real-world 1D, 2D, and 3D datasets. Different strategies of achieving the identity and orthogonality constraints for DeepJIVE were explored, resulting in three viable loss functions. We found that DeepJIVE can successfully uncover joint and individual variations of multimodal datasets. Our application of DeepJIVE to the Alzheimer's Disease Neuroimaging Initiative (ADNI) also identified biologically plausible covariation patterns between the amyloid positron emission tomography (PET) and magnetic resonance (MR) images. In conclusion, the proposed DeepJIVE can be a useful tool for multimodal data analysis.

## 1. Introduction

Large multimodal datasets have become widely available in many fields, including biomedical research. Data analysis methods that examine the relationship between and within each modality of such multimodal data can provide key insights to different aspects of the subject that each modality captures. Indeed, several conventional statistical methods have been developed for multimodal data that are in tabular form, such as Consensus Principal Component Analysis,[1] Multiple Canonical Correlation Analysis (mCCA),[2] and Joint and Individual Variance Explained (JIVE).[3]

---

[*] Data used in preparation of this article were obtained from the Alzheimer's Disease Neuroimaging Initiative (ADNI) database (adni.loni.usc.edu). As such, the investigators within the ADNI contributed to the design and implementation of ADNI and/or provided data but did not participate in analysis or writing of this report. A complete listing of ADNI investigators can be found at: http://adni.loni.usc.edu/wp-content/uploads/how_to_apply/ADNI_Acknowledgement_List.pdf

These methods have been successfully applied in studying multimodal data such as measures of biological chemicals including proteomics and metabolomes,[4-8] meteorological data,[9] and neurological images,[10,11] and some have found applications in facial image analysis.[12,13] However, these conventional statistical methods typically require dimensionality reduction of high-dimensional data, such as images, to lower dimensionality, typically less than a few hundred. However, dimensionality reduction may entail the loss of vital information inherent in the original high-dimensional data. For example, modern 3D structural brain magnetic resonance (MR) images typically have several million voxels, and preprocessing is typically performed to reduce the data to calculate regional volume of a couple hundred brain regions. Conventional statistical methods cannot handle the imaging data directly due to computational intractability and nonlinear dependency between different voxels in the original images. The advance of AI technologies has demonstrated great success in directly handling high-dimensional data, such as images. For example, convolutional neural networks have demonstrated great success in classifying images[14-16] and in image segmentation using the U-Net architecture[17] and its derivatives among other tasks. In this work, we propose a deep-learning based approach as a conceptual extension to JIVE to directly handle images and uncover higher-order joint and individual structures in the high-dimensional data. We termed this approach, DeepJIVE. We will first revisit the theory of JIVE, followed by a description of our deep learning-based extension. We evaluate and test the performance of DeepJIVE on 1D simulation, 2D simulation, and apply the method in a dataset from the Alzheimer's Disease Neuroimaging Initiative (ADNI) to undercover joint and individual variations between MR and positron emission tomography images.

## 2. Theory

### 2.1 JIVE

JIVE posits that each data modality $X_k$, indexed by $k \in \{1, 2, ..., K\}$, can be decomposed as a linear combination of three separate components: the joint component, $J_k$, representing variation explained by underlying factors common to all data modalities; the individual component, $S_k$, representing variation explained by underlying factors unique to each data modality; and the noise component, $e_k$, representing variation due to measurement error (Eq. 1) [3]

$$X_k = J_k + S_k + \varepsilon_k \qquad [Eq. 1]$$

where $X_k$ is a $p_k \times n$ matrix of $p_k$ variables and $n$ subjects and, using row() to denote the row space or score space of a matrix, row($J_k$) = row($J_{j \neq k}$) and row($J_k$) ^ row($S_k$) for all $k$. This is illustrated in Figure 1 with examples from a dataset used in Section 2.4.2. The images represent the loadings of different components. Here, for illustration purposes and similar to what is done in Lock[3], we make the loadings the same for the joint component of each different data modality, and the scores are randomly generated from a Gaussian distribution. However, this assumption is not made in our models. The first set of images represents the loadings of $J_k$ for every $X_k$, and the second represents S, which are unique for each $X_k$; these combine to form X, shown on the right. Not all $J_k$ need to be identical as they are in this example, but all $J_k$ do need to be generated from the same underlying process, i.e. have the same score space.

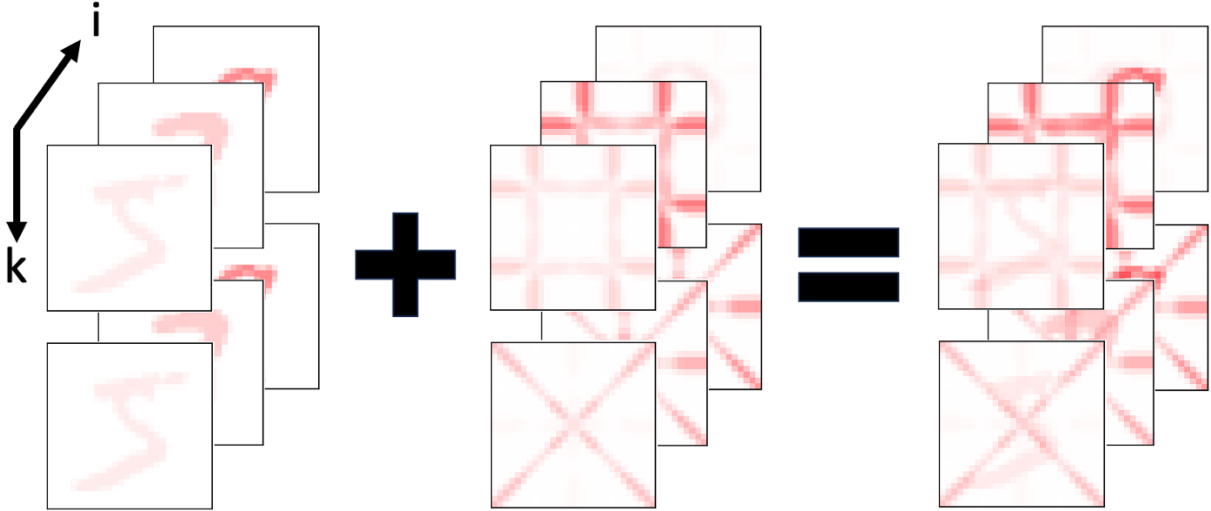

**Figure 1.** An illustration of a dataset used in Section 2.4.2. The top and bottom images of each row represent different data types, $k \in \{1, 2, … K\}$, for each sample, $i \in \{1, 2, …, n\}$. The joint structure, $J_k$, on the left is the same for all k in this data set, but it does not need to be identical as long as they are always generated from the same underlying process.

The joint and individual components can be parameterized as the product of a matrix of loading vectors and a matrix of scores (Eq. 2-4), where $\Lambda_J$ is the $r_J \times n$ score matrix common to all data modalities and is not indexed, $U_{J_k}$ is the $p_k \times r_J$ loading matrix for the joint component in the k-th data modality, $\Lambda_{S_k}$ is the $r_{S_k} \times n$ score matrix the individual component of the k-th data modality, and $U_{S_k}$ is the $p_k \times r_{S_k}$ loading matrix for the individual component of the k-th data modality.

$$J_k = U_{J_k}\Lambda_J \qquad [Eq. 2]$$

$$S_k = U_{S_k}\Lambda_{S_k} \qquad [Eq. 3]$$

$$X_k = U_{J_k}\Lambda_J + U_{S_k}\Lambda_{S_k} + \varepsilon_k \qquad [Eq. 4]$$

The seminal paper by Lock et. al.[3] solved for these structures using an iterative process where the rank $r_J$ or $r_{S_k}$ principal component analysis (PCA) decomposition of $J_k$ or $S_k$, respectively, was used to find the next estimate of the other structure. This iterative process is time consuming, so others have developed other methods of determining these structures more directly, such as Angular JIVE (AJIVE) and Canonical JIVE

(CJIVE).[3,18,19] These methods still rely heavily on PCA as it is an effective method of extracting common information from many variables, and it also helps satisfy the main constraints of JIVE: that the noise structure be removed from the joint and individual structures ($r_J < \sum_k^K r_{X_k}$ and $r_{S_k} < r_{X_k}$), known as the rank constraint; and that no information should be shared between $J_i$ and $S_i$ (the rows of $J_k$ and $S_k$ must be orthogonal), known as the orthogonality constraint. While these constraints make PCA useful for solving JIVE, PCA only finds linear transformations and relationships in the data, which can somewhat limit its applicability to high-dimensional data such as images. Our work aims to extend beyond this linearity assumption and have the capacity of working directly on high-dimensional data such as images.

## 2.2 Relationship between PCA and autoencoders

Autoencoders, like PCA, are a dimensionality reduction method as they summarize complex, multivariate data using a smaller set of variables referred to as the latent space. Autoencoders are composed of two neural networks: the encoder, $f^E(x_i)$, that transforms the original data, $x_i \in X$, into the latent space, $\lambda_i \in \Lambda$; and the decoder, $f^D(\lambda_i)$, that transforms the latent space back into the original data space to produce $x_i$'s reconstruction, $\hat{x}_i \in X$.

$$\lambda_i = f^E(x_i) \qquad [Eq.\,5]$$

$$\hat{x}_i = f^D(\lambda_i) \qquad [Eq.\,6]$$

$$L = \sum_i^n (x_i - \hat{x}_i)^2 = \sum_i^n \left(x_i - f^D\big(f^E(x_i)\big)\right)^2 \qquad [Eq.\,7]$$

Because $\Lambda$ is $p_\Lambda \times n$, and $p_\Lambda < p_X$, the autoencoder learns a low-dimensional representation of the data and a pair of transformations to and from this representation by training the networks to minimize the reconstruction error defined in Eq. 7. Direct comparisons have been drawn between autoencoders and PCA as they seek to minimize the same objective function. In particular, it has been found that a single-layer linear autoencoder will identify the same subspace as PCA[20,21]. However, autoencoders have an advantage over PCA as they are capable of learning non-linear functions with the use of non-linear activation functions and multiple layers including convolutional layers. We hypothesize that autoencoders can be viewed as a non-linear extension of PCA and should therefore be used as building blocks for performing linear or non-linear versions of JIVE that is capable of handling high-dimensional data.

## 2.3 DeepJIVE

### 2.3.1 Theory

In JIVE, the joint and individual components $J_k$ and $S_k$ are linear function of the joint and individual scores $\Lambda_J$ and $\Lambda_{S_k}$, respectively (Eqs. 2 and 3). Analogous to the linear equations Eqs. 2 and 3 in JIVE, DeepJIVE seeks to learn a non-linear extension of JIVE so that that $J_k$ and $S_k$ are nonlinear functions of $\Lambda_J$ and $\Lambda_{S_k}$:

$$J_k = f_{J_k}^D(\Lambda_J), \qquad [Eq.\,8]$$

$$S_k = f_{S_k}^D(\Lambda_{S_k}), \qquad [Eq.\,9]$$

where $f_{J_i}^D()$ and $f_{S_i}^D()$ are realized using neural networks parameterized by their learnable weights. In reverse, the joint and individual scores, or in machine learning terminology, latent variables, are calculated from each data modality $X_k$ using learned functions $f_{J_k}^E()$ and $f_{S_k}^E()$, also parameterized by neural networks:

$$\Lambda_J = f_{J_k}^E(X_k) \qquad [Eq.\,10]$$

$$\Lambda_{S_k} = f_{S_k}^E(X_k) \qquad [Eq.\,11]$$

Since $\Lambda_J$ is the same for all data modalities, $f_{J_k}^E(X_k) = f_{J_{j \neq k}}^E(X_{j \neq k})$ for all k and j. Similar to conventional JIVE, we also would like to have the property of orthogonality between $\Lambda_J$ and $\Lambda_{S_k}$.

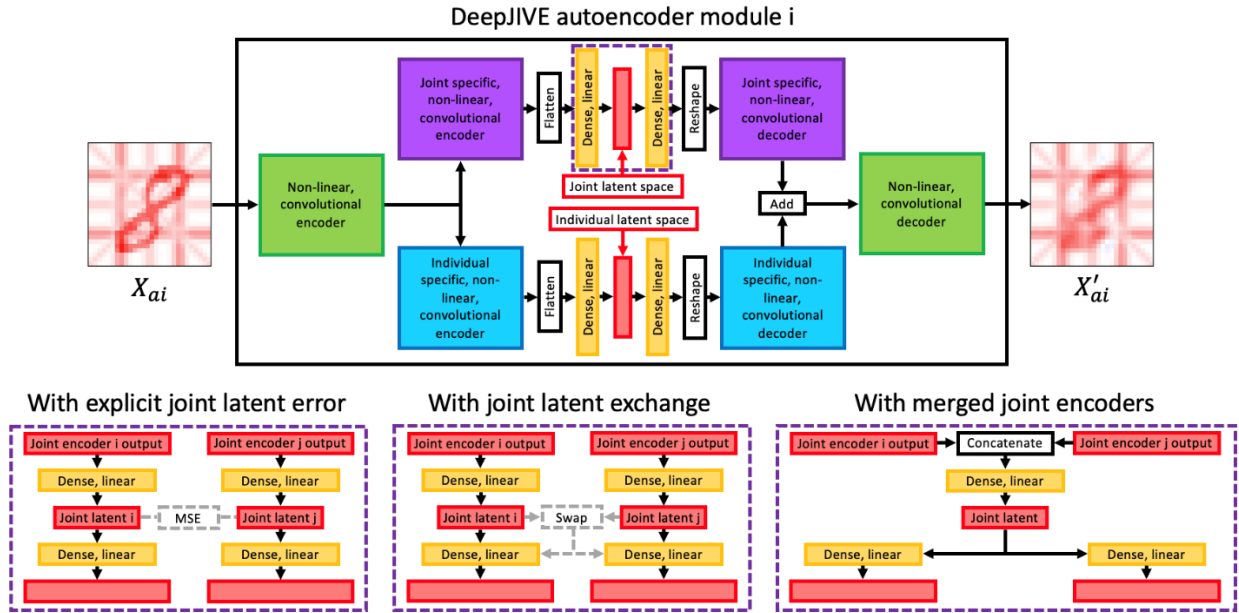

**Figure 2.** (Top) The basic layout of the DeepJIVE architecture. The core architecture of a DeepJIVE network is a set of parallel pairs of autoencoders. One from each pair represents the joint structure while the other represents the individual structure, and there are as many of these pairs as there are data types ($k \in K$). While some layers of the encoder and decoder can be shared by both, the input to the joint- and individual specific encoders is the same, while the outputs of the joint- and individual-specific decoders are added together. (Bottom) Three variants of the joint-specific autoencoders. The first two use the same architecture as shown in the top diagram, but the latent space is used differently during training (shown by the gray lines). The third variant represents a different architecture, where the outputs of each joint-specific, non-linear encoder are flattened and concatenated to be fed into a singular encoder.

### 2.3.2 Architecture, loss function, and training

The key component to the DeepJIVE architecture is a pair of parallel autoencoders shown in the top box of Figure 2, one representing the joint structure (Eq. 8 and 10) and the other representing the individual structure (Eq. 9 and 11). The input to each of these is the same, although it need not be the raw sample as some non-linear, convolutional layers can be shared between them, if desired, although the outputs of their respective decoders are added together, same in the conventional JIVE formation represented in Eq. 1. Additionally, each data type being analyzed as part of the same JIVE analysis has their own pair of autoencoders, so there should be a total of K pairs for K data types. Because there are as many joint latent vectors, $\Lambda_{J_k}$, as there are data types, the DeepJIVE network requires the constraint of producing identical values from each joint encoder, which is implicit in the conventional JIVE formulation. In DeepJIVE, this constraint is satisfied primarily by using different loss functions during training, or different architectures for the final layer of the joint-specific encoder.

This paper examines three strategies to satisfy this identity constraint, the first of which involves additional loss terms to account for the latent space, the second of which involves multiple reconstruction terms, and the third of which involves a modified architecture to achieve the identity constraint. The first strategy that uses the default architecture shown in Figure 2 minimizes the mean squared error (MSE) between the input, $X_k$, and its reconstruction from the output, $f_{J_k}^D(\Lambda_{J_k}) + f_{S_k}^D(\Lambda_{S_k})$, of each pair of autoencoders as well as the MSE of the joint latent vectors, $\Lambda_{J_k} = f_{J_k}^E(X_k)$, between all data types.

$$L_{explicit} = \sum_{k=1}^{K} \left\| X_k - \hat{X}_k \right\|_2 + \sum_{k=1}^{K-1} \sum_{j=k+1}^{K} \left\| \Lambda_{J_k} - \Lambda_{J_j} \right\|_2$$

$$= \sum_{k=1}^{K} \left\| X_k - \left( f_{J_k}^D \left( f_{J_k}^E(X_k) \right) + f_{S_k}^D \left( f_{S_k}^E(X_k) \right) \right) \right\|_2 + \sum_{k=1}^{K-1} \sum_{j=k+1}^{K} \left\| f_{J_k}^E(X_k) - f_{J_j}^E(X_j) \right\|_2 \quad [Eq. 12]$$

This is represented by the box in Figure 2 labeled "With explicit joint latent error" and will hereafter be referred to as "explicit" networks, because the latent space vector is explicitly enumerated in the loss equation. The second strategy does not explicitly train from the latent space vectors, but it does so implicitly because not only does it train from the MSE between the input and the output produced by each data modalities' own joint score vector, $f_{J_k}^D(\Lambda_{J_k})$, but each other data modalities', $f_{J_k}^D(\Lambda_{J_{j \neq k}})$, as well.

$$L_{exchange} = \sum_k^K \sum_j^K \left\| X_k - \left( f_{J_k}^D \left( f_{J_j}^E(X_j) \right) + f_{S_k}^D \left( f_{S_k}^E(X_k) \right) \right) \right\|_2 \quad [Eq.\,13]$$

This produces additional loss terms from the second summation operation over j. This satisfies the identity constraint by ensuring that all latent vectors produced by the joint encoders are functionally identical to one another, and DeepJIVE networks trained using this method will be referred to as "exchange" networks because the joint latent vectors are exchanged between each joint decoder during training. The final strategy does not use the default architecture shown in the top box of Figure 2 but instead produces a single joint latent vector from the combined outputs of the joint specific non-linear encoders. The outputs of joint specific encoders, $f_{J_k}^E(X_k)$, are flattened and concatenated into a single vector, then fed into a single dense layer to produce the joint latent vector. Because this results in a single set of joint scores, the identity constraint is inherently satisfied, and the network is trained on simple reconstruction error.

$$L_{merged} = \sum_k^K \left\| X_k - \left( f_{J_k}^D(\Lambda_J) + f_{S_k}^D \left( f_{S_k}^E(X_k) \right) \right) \right\|_2 \quad [Eq.\,14]$$

$$\Lambda_J = f_J^E\left( \left[ f_{J_0}^E(X_0), \ldots, f_{J_K}^E(X_K) \right] \right)$$

This architecture corresponds to the "with merged joint encoders" box in Figure 2 because it involves merging the output of the joint encoders into a single vector, and these types of DeepJIVE networks will thus be referred to as "merged" networks. An additional consideration for this network is that it requires all data types to produce the true joint vector while the other types can produce the joint latent vector with only one data type as replacing any $f_{J_k}^E(X_k)$ with $\vec{0}$ will affect the outcome of $f_J^D\left( \left[ f_{J_0}^E(X_0), \ldots, f_{J_K}^E(X_K) \right] \right)$.

### 2.3.3 Orthogonality constraint

The latent variables of a linear autoencoder will trend towards orthogonality, but the landscape of the loss function contains many saddle points which can prevent the network from achieving complete orthogonality.[21] DeepJIVE networks are no more immune from this than other autoencoders, but they require orthogonality more than other autoencoders, particularly between the joint, $\Lambda_J$, and individual, $\Lambda_{S_k}$, latent space variables. This issue can be overcome using additional terms and mechanisms to smooth out the loss surface, such as using a regression network to learn the relationships between $\Lambda_J$ and $\Lambda_{S_k}$ during training and then removing the variance from each $\Lambda_{S_k}$ explained by $\Lambda_J$.

$$L_{regression} = \left\| f_{S_k}^E(X_k) - f_k^R\left( f_J^E(X_k) \right) \right\|_2 \quad [Eq.\,15]$$

These regression networks, $f_k^R(\Lambda_J)$, can be as simple as a single linear layer and are trained by minimizing the mean squared error between its output and $\Lambda_{S_k}$. While they could take $\Lambda_{S_k}$ as the input and train off the error between $\Lambda_J$ and $f_k^R(\Lambda_{S_k})$, it is recommended that $\Lambda_J$ be used as the input since there will need to be one regression network for each data modality, so providing k – 1 additional, possibly contradictory gradients may inhibit learning. The additional term added to the loss equation of the DeepJIVE network is essentially the magnitude of the regression networks' output.

$$L_{ortho} = \left\| f_k^R\left(f_J^E(X_k)\right) \right\|_2 = \left\| f_{S_k}^E(X_k) - \Lambda_{S_k}^* \right\|_2 \quad [Eq.\,16]$$

$$\Lambda_{S_k}^* = f_{S_k}^E(X_k) - f_k^R\left(f_J^E(X_k)\right) \quad [Eq.\,17]$$

While the two versions of Equation 16 are equivalent, it is important to frame the loss equation in terms of $f_{S_k}^E(X_k)$ rather than as a function of $f_J^E(X_k)$ as there are two ways to minimize $f_k^R(\Lambda_J)$, remove any dependence between $\Lambda_{S_k}$ and $\Lambda_J$ or minimize $\Lambda_J$, the latter of which is to be avoided. This turns the full loss equation of the DeepJIVE network into:

$$L_{DeepJIVE} = L_{arch} + \gamma L_{ortho} \quad [Eq.\,18]$$

where $L_{arch}$ is either $L_{explicit}$, $L_{exchange}$, or $L_{merged}$ from Equations 12, 13, or 14, respectively, and $\gamma$ is a weighting factor. $\gamma$ should generally be kept low as the orthogonality constraint should not prevent the network from leaving its initial orthogonal state. Similarly, the learning rate for the regression network should be higher than that of the DeepJIVE network as it needs to quickly react to any emerging dependencies in order to provide accurate gradients.

### 2.3.4 Rank selection

Choosing the appropriate $r_J$ and $r_{S_k}$ is important for JIVE as choosing an $r_J$ that is too low for the respective $r_{S_k}$ can cause joint information to be captured by the individual components, but choosing an $r_J$ that is too high can cause the joint structure to capture individual information. The method for choosing $r_J$ used for AJIVE can also be applied to DeepJIVE,[18] although the initial determination of the initial score matrices for each data type is different due to the deep learning nature of DeepJIVE. Because the decomposition of each datatype into $\Lambda$ is dependent on the architecture of $f^E(X_k)$, separate autoencoders without separate latent spaces for $\Lambda_J$ and $\Lambda_{S_k}$ need to be trained for each datatype, each using the same train:test split in order to be able to sync the samples from each common subject in subsequent analysis. The architectures of these separate autoencoders should be the same as those intended to comprise the DeepJIVE network, except for the number latent space components, which should be larger. As illustrated by Plaut, the loading vectors of the PCA decomposition of a dataset are the same as the loading vectors of a PCA decomposition of the decoder weight matrix of a linear autoencoder trained on that dataset because an autoencoder spans the same subspace as PCA, assuming it is trained sufficiently to minimize reconstruction error.[20] Similarly, the row space of a PCA decomposition can be extracted from a PCA decomposition of the latent space matrix for the training set as two matrix transformations that produce the same result and span the same column space must have the same row space. Once the score spaces for each datatype's separate autoencoders have been extracted by PCA, the total rank of each datatype, $r_k$, can be taken to be the number of components where the reconstruction error vs number of components curve or its derivative crosses a desired, arbitrary threshold. Once the total rank of each datatype, $r_k$, are selected, $r_J$ selection can proceed as in AJIVE: concatenate the $r_k \times n$ score matrices along their variable dimension and perform PCA on the concatenated matrix. $r_J$ is determined by the number of components that have a singular value above a given threshold. This threshold is determined by simulation, generating matrices of random data with the same dimensions and distributions as the real data, concatenating them, and finding the highest singular value for these simulated datasets. By repeating this simulation process a sufficient number of times, one can build up a distribution of the highest singular value that might have been generated from your data by random chance, so any component from the real dataset with a singular value greater than the 95% of the simulated distribution can be said to have not occurred randomly with 95% confidence.

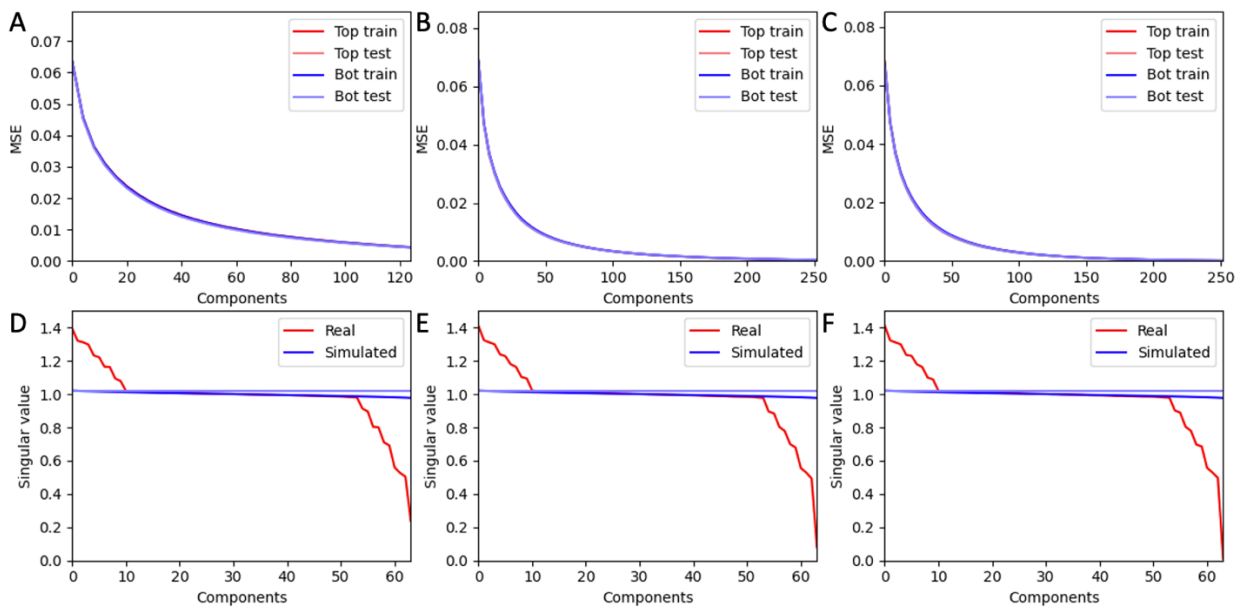

**Figure 3.** Component selection criteria diagrams for the merged networks used in the MNIST paired dataset (Section 2.4.2). A, B, and C show reconstruction MSE curves for a given number of principal components. D, E, and F show the singular values from the SVD of the concatenated score matrices corresponding to the networks used to produce the figures above them compared against the average of 100 simulated random matrices of the same number of samples and variables. A and D are produced from linear networks, B and C are produced by networks with a single convolutional layer, and C and F correspond to networks with two convolutional layers.

## 2.4 Experiments

The three DeepJIVE variants detailed in the previous section, merged networks, exchange networks, and explicit networks, were evaluated using four separate datasets, each of which were intended to validate a specific capability of DeepJIVE. The first dataset is one-dimensional and seeks to directly establish the capability of DeepJIVE to replicate conventional JIVE in a linear and low-dimensional setting. The second two datasets are two-dimensional that evaluate the effects of convolutional and non-linear layers on performing JIVE. The fourth seeks to provide a practical demonstration of DeepJIVE using real-world data. The networks used for each dataset are described in the following sections, but a summary of each is given in Table 1.

**Table 1.** Details for the DeepJIVE architectures used in the experiments discussed in this paper.

| | Experiment (Section) | 1D synthetic (3.1) | MNIST overlaid (3.2.1) | MNIST paired (3.2.2) | ADNI MRI-PET (3.3) |
|---|---|---|---|---|---|
| Data | Samples (train:test) | 100 (90:10) | 70,000 (60,000:10,000) | 70,000 (60,000:10,000) | 821 (727:94) |
| | Size (type 1), (type 2) | (100); (100) | (28, 28); (28, 28) | (28, 28); (28, 28) | (60, 72, 60); (48, 56, 48) |
| Layers* | Data type 1 (kernels, convolution, stride, activation) | -- | 44; (3, 3); (1, 1); ReLU<br>44; (3, 3); (1, 1); ReLU<br>44; (3, 3); (1, 1); ReLU<br>44; (3, 3); (1, 1); ReLU<br>44; (3, 3); (1, 1); ReLU | 32; (3, 3); (2, 2); ReLU<br>64; (3, 3); (2, 2); ReLU | 16; (3, 3, 3); (1, 1, 1); ReLU<br>16; (3, 3, 3); (2, 2, 2); ReLU<br>32; (3, 3, 3); (1, 1, 1); ReLU<br>32; (3, 3, 3); (2, 2, 2); ReLU |
| | Data type 2 (kernels, convolution, stride, activation) | -- | 44; (3, 3); (1, 1); ReLU<br>44; (3, 3); (1, 1); ReLU<br>44; (3, 3); (1, 1); ReLU<br>44; (3, 3); (1, 1); ReLU<br>44; (3, 3); (1, 1); ReLU | 32; (3, 3); (2, 2); ReLU<br>64; (3, 3); (2, 2); ReLU | 16; (3, 3, 3); (1, 1, 1); ReLU<br>16; (3, 3, 3); (2, 2, 2); ReLU<br>32; (3, 3, 3); (1, 1, 1); ReLU<br>32; (3, 3, 3); (2, 2, 2); ReLU |
| Latent | Joint | 1 | 10 | 10 | 34 |
| | Data type 1 | 1 | 2 | 26 | 26 |
| | Data type 2 | 1 | 2 | 26 | 142 |

*For experiments that use multiple networks with different numbers of layers, the layers listed describe the architecture of the largest network used in the experiment. The networks with fewer layers have the same architecture up to the number of layers given for that network.

### 2.4.1 1D synthetic data

The first test reported in this paper is on two sets of 100 vectors each containing 100 variables designed to emulate the initial test in the seminal paper on JIVE.[3] The value of each variable was determined by

$$x_{ijk} = \alpha_{ij} + \beta_{ik} + \varepsilon_{ijk} \begin{cases} \alpha \in [-2, 2] \text{ if } j \leq 50; \{0\} \text{ if } j > 50 \\ \beta \in \{-2, -1, 0, 1, 2\} \\ \varepsilon \in [-0.5, 0.5] \end{cases} \quad [Eq. 19]$$

where $i$ represents the sample, or columns of the matrices in Figure 6, $j$ represents the variables of each vector, $j \in p_k$, and $k$ represents the vector set, or the top or bottom matrix. $\alpha_{ij}$, $\beta_{ik}$, and $\varepsilon_{ijk}$ are uniformly distributed variables, with $\alpha_{ij}$ and $\varepsilon_{ijk}$ being continuous variables and $\beta_{ik}$ being discrete. $\alpha_{ij}$ represents the joint structure as it is shared between both vectors while $\beta_{ik}$ represents the individual structure. The representation of this data in Figure 6 is sorted by the individual structure of the first vector for the sake of visualization. The three DeepJIVE networks used for this experiment, merged, exchange, and explicit, only contained a single linear layer.

**2.4.2 2D synthetic data**

The second test seeks to demonstrate how DeepJIVE can expand the applicability of JIVE by expanding the data from one dimension to two dimensions and incorporating convolutional, nonlinear layers to the network. Two separate datasets were prepared based on the MNIST handwritten image dataset,[22] both consisting of two images, hereafter referred to as the top and bottom images. The first 2D dataset, hereafter referred to as the MNIST overlaid dataset, was designed to directly translate the 1D dataset into two dimensions. The MNIST overlaid dataset was constructed by adding the same pattern, representing the joint structure, to two unique patterns generated from the same distribution, representing the individual structure, as described by Equation 20.

$$x_{ik} = .5 \sum_{j=0}^{9} \alpha_{ij} m_{jk} + .5(\beta_{ik1} b_{k1} + \beta_{ik2} b_{k2}) \quad [Eq. 20]$$

In this case, $m_{jk}$ is one of ten images, one for each digit, from the MNIST dataset scaled to have intensities ranging from 0 to 1, and b$_{k1}$ and b$_{k2}$ are two of the synthetic images from Figure 4. $m_{jk}$ and b$_k$ correspond to $U_{Jk}$ and $U_{S_k}$, the loading vectors from Equation 19, while their corresponding coefficients, $\alpha_{ij}$ for $m_{jk}$ and $\beta_{ik}$ for b$_k$, correspond to $\Lambda_J$ and $\Lambda_{S_k}$, the score matrices from Equation 19.[22] $\beta_{ik}$ was uniformly randomly distributed between 0 and 1, but only one $\alpha_{ij}$ per sample i was non-zero, so if $\alpha_{ip} \neq 0$, then $\alpha_{ij \neq p} = 0$. The index j chosen to be non-zero for $\alpha_{ij}$ was chosen uniformly randomly, and the non-zero value was uniformly randomly distributed between 0 and 1. The first term of Equation 20, $\sum_{j=0}^{9} \alpha_{ij} m_{ij}$, represents the joint structure and corresponds to $\alpha_{ij}$ in Equation 19; the second term, $\beta_{ik1} b_{k1} + \beta_{ik2} b_{k2}$, represents the individual structure and corresponds to b$_{ik}$ in Equation 19. It is noteworthy that in this simulation the loadings, $m_{jk}$, were chosen to be the same between the two data types; this decision, along with the decision to make only one score variable $\alpha_{ij}$ non-zero per sample, was made for visualization purposes. In practice, the loadings are typically different between different data types, and only the joint scores are the same between different data types. Each DeepJIVE architecture used for this dataset the same number of 3x3 convolutional kernels, and networks up to five layers were tested. Each layer had 44 kernels and used ReLU activation functions; a linear network served as a baseline with which other models were compared.

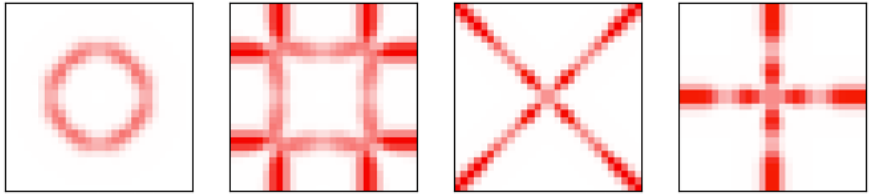

**Figure 4.** The four patterns corresponding to $b_{11}$, $b_{12}$, $b_{21}$, and $b_{22}$ from left to right in Equation 20.

The second dataset, hereafter referred to as the MNIST paired dataset, was constructed by pairing each image of the MNIST dataset with another image of the same digit; both image sets used all images from the MNIST dataset, but the images were paired such that the same image was never paired with itself. Each DeepJIVE architecture used for this dataset used shared non-linear layers for each data type, and networks up to two layers were tested. The first convolutional layer, if present, had 16 3x3 convolutional kernels, had a stride length of 2x2, and used ReLU activation functions while the second, if present, had 32 kernels of the same size, stride length, and activation function; a linear network served as a baseline with which other models were compared.

### 2.4.3 ADNI dataset and image pre-processing

The final test seeks to demonstrate how DeepJIVE networks can be used to analyze real-world datasets. This test used a dataset consisting of T1w MRI images and amyloid (florbetapir, AV45) PET images obtained from the Alzheimer's Disease Neuroimaging Initiative (ADNI), including cognitively normal (CN) participants, participants with mild cognitive impairment (MCI), and participants with early Alzheimer's disease (AD). For up-to-date information, see www.adni-info.org. T1w MR and amyloid PET images were paired based on collection date, grouping the two temporally closest images taken no more than six months apart for each subject. Only the earliest image pair from each subject was used for network training, resulting in 821 image pairs split into training and validation datasets at a 9:1 ratio. While not used for training, the other 1117 pairs were included in datasets for analysis. Raw T1w images and amyloid PET images were registered to the MNI space using SPM12 (version 7771),[23] and the T1w images were segmented using SPM12 to obtain gray matter tissue probability maps (TPM) in the MNI space. The intensity of the PET images were normalized to the average intensity of their cerebellum as defined by the atlas provided by the Centiloid Project,[24] and a brain mask from the FMIRB Software Library (FSL) was used to remove the skull and non-cortical tissue regions.[25] The TPM and intensity-normalized PET images were then rescaled to have a voxel size of 3x3x3 mm³ by average pooling to reduce memory requirements.

## 3. Results

### 3.1 1D simulation

To demonstrate the effectiveness of DeepJIVE, the three different strategies for implementing DeepJIVE described in the previous section were trained to replicate a test performed in the seminal paper on JIVE by Lock, Hoadley, and Marron.[3] This particular example is particularly difficult for DeepJIVE to solve as there are many different minima in which the network can become trapped. Consider another dataset similar to Equation 19:

$$x_{ijk} = \alpha'_{ij} + \beta'_{ik} + \varepsilon_{ijk} \begin{cases} \alpha'_{ij} \in [\gamma - 2, \gamma + 2] \text{ if } k \leq 50, \gamma \text{ if } k > 50 \\ \beta'_{ik} \in \{-\gamma - 2, -\gamma - 1, -\gamma, -\gamma + 1, -\gamma + 2\} \\ \varepsilon \in [-0.5, 0.5] \end{cases} \quad [Eq. 21]$$

The dataset constructed using Equation 21 is essentially the same as that using Equation 19, except the distributions of the constituent variables $\alpha'_{ij}$ and $\beta'_{ik}$ are offset from $\alpha_{ij}$ and $\beta_{ik}$ in opposite directions by a constant magnitude, $\gamma$. While the ground truths of the joint and individual structures for the data generated by Equations 19 and 21 could differ by orders of magnitude, both will add together to form the same x, and any information about $\gamma$ will be lost. Equation 21 illustrates that for Equation 19, there is not necessarily a single ground truth with a large number of local minima but an infinite number of ground truths. Additionally, because $\beta_{ik}$ is constant for all j, any $U_S$ or $f_S^D$ that parameterizes $\beta_{ik}$ will be able to compensate for a $U_J$ or $f_J^D$ that encodes a non-zero $\gamma$ simply by scaling $\Lambda_J$. This is demonstrated in Figure 5A, showing the weights and latent space variables of DeepJIVE networks trained on the 1D dataset defined by Equation 19. As can be seen in Figure 5C, there is much less correlation between the joint and individual components of the networks trained with the $L_{ortho}$ term than those without by virtue of the shallower slopes of the lines that appear in the plots of $f_{S_k}^E(x_{ik})$ vs. $f_J^E(x_{ik})$. The reason for these slopes is seen in Figure 5A with the plots of the weight vectors for $f_J^D$. The joint weight vectors for the networks trained with $L_{ortho}$ tend have near 0 values for one half of the vector while the joint weight vectors of the networks trained without $L_{ortho}$ tend to be centered on 0. This centering acts as a non-zero $\gamma$, but since $f_{S_k}^D$ is a flat line, it could either learn this non-zero $\gamma$ with a bias or by scaling $\Lambda_{S_k}$. Based on the slopes in Figure 5C, it does this at least in part using the latter method, creating an correlation between the joint and individual structures. The networks trained with $L_{ortho}$ do not have this dependence and are thus able to produce better estimates of J and S as shown in Figure 6.

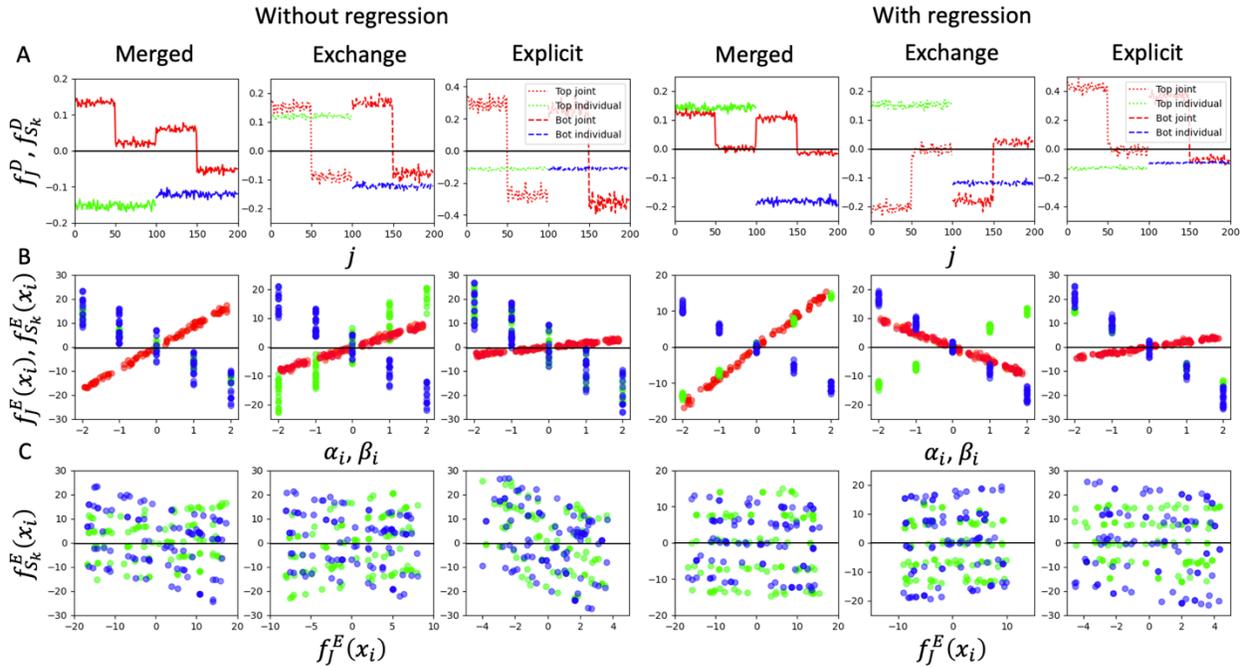

**Figure 5.** A) Weight vectors of the joint (red) and individual (green, blue) decoders trained with (right) and without (left) a regression network to enforce orthogonality between joint and individual components. The networks trained with a regression network have weight vectors closer to $\alpha_{ij}$ from Equation 19 with one half being non-zero and the other being near zero while the networks trained without a regression network tend to have weight vectors centered closer to 0. B) Latent space values produced by the networks plotted against the original $\alpha$ and $\beta$ values for each sample. A greater vertical spread in the individual latent space values indicates a stronger correlation between joint and individual components as the individual component compensates for an offset in the joint component. C) Individual latent space values plotted against their corresponding joint latent space values. While not perfectly flat, the networks trained with a regression network show smaller slopes than those without, demonstrating the ability for the regression network to enforce orthogonality.

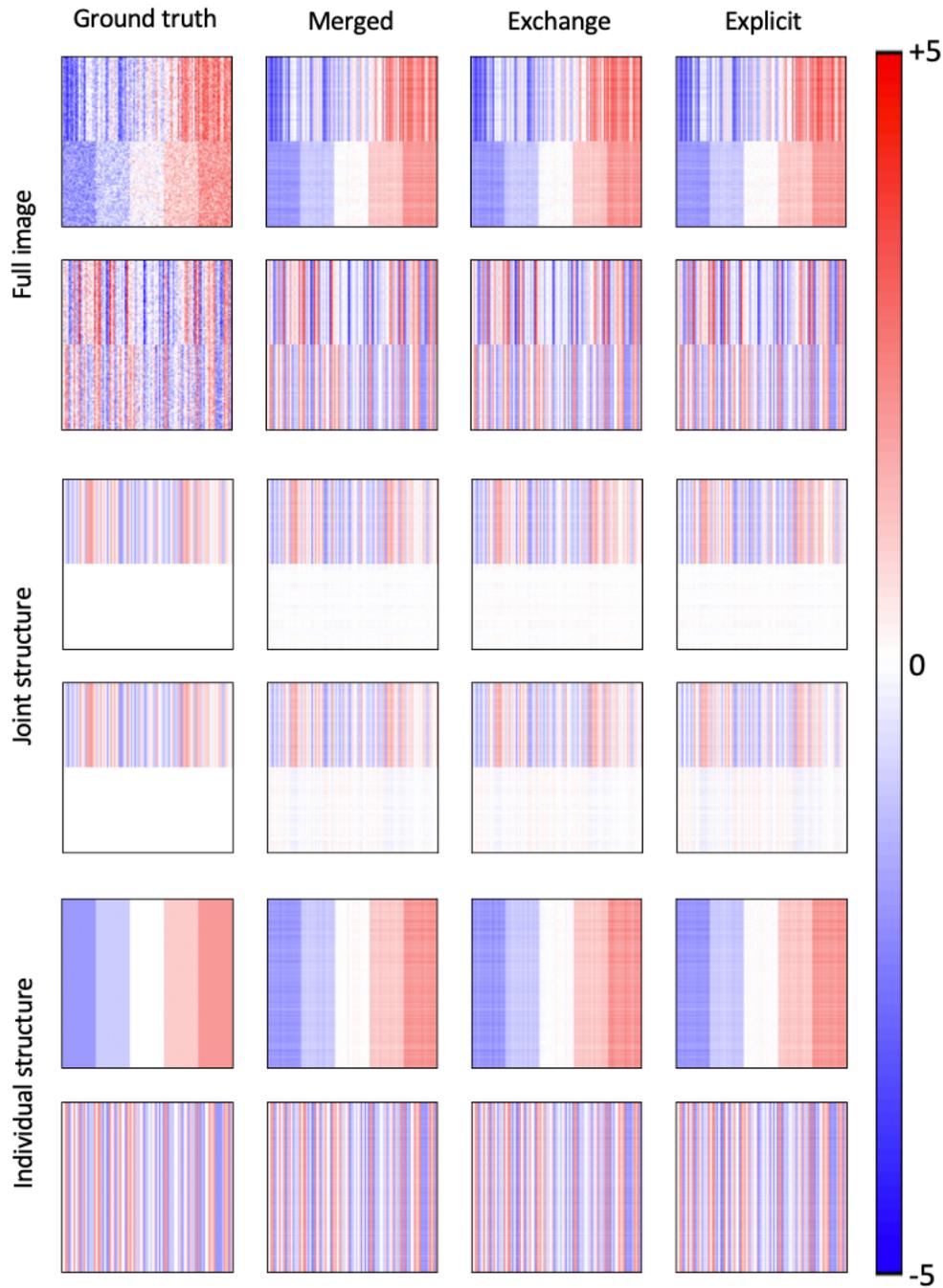

**Figure 6.** Joint structure, individual structure, and full data reconstructions from explicit, exchange, and merged networks trained on the 1D dataset defined by Equation 19.

### 3.2 2D simulation with convolutional layers

#### 3.2.1 MNIST overlaid dataset

The MNIST overlaid dataset uses the philosophy for the construction of the 1D dataset but expands on its complexity by increasing the number of dimensions of the data and the number of components used to construct it for both the joint and individual structures. Figure 7 shows the synthesized output of merged networks trained using different numbers of convolutional layers without a regression network using specific groups of latent variables, and even without an orthogonality enforcement mechanism, DeepJIVE networks naturally separate the information related to the common digit and the overlaid patterns between the joint and individual variables, respectively. While the number of convolutional layers did not have much impact on the overall reconstruction, the amount of bleed over between the joint and individual structure generally reduces with the number of convolutional layers. The quality of the joint structure, in comparison to the constituent image, also tends to degrade with the number of convolutional layers, but the more non-linear a function becomes, the more impact a missing parameter will have on the overall output, so removing the joint or individual latent variables may still have some impact on the reconstruction of the other.

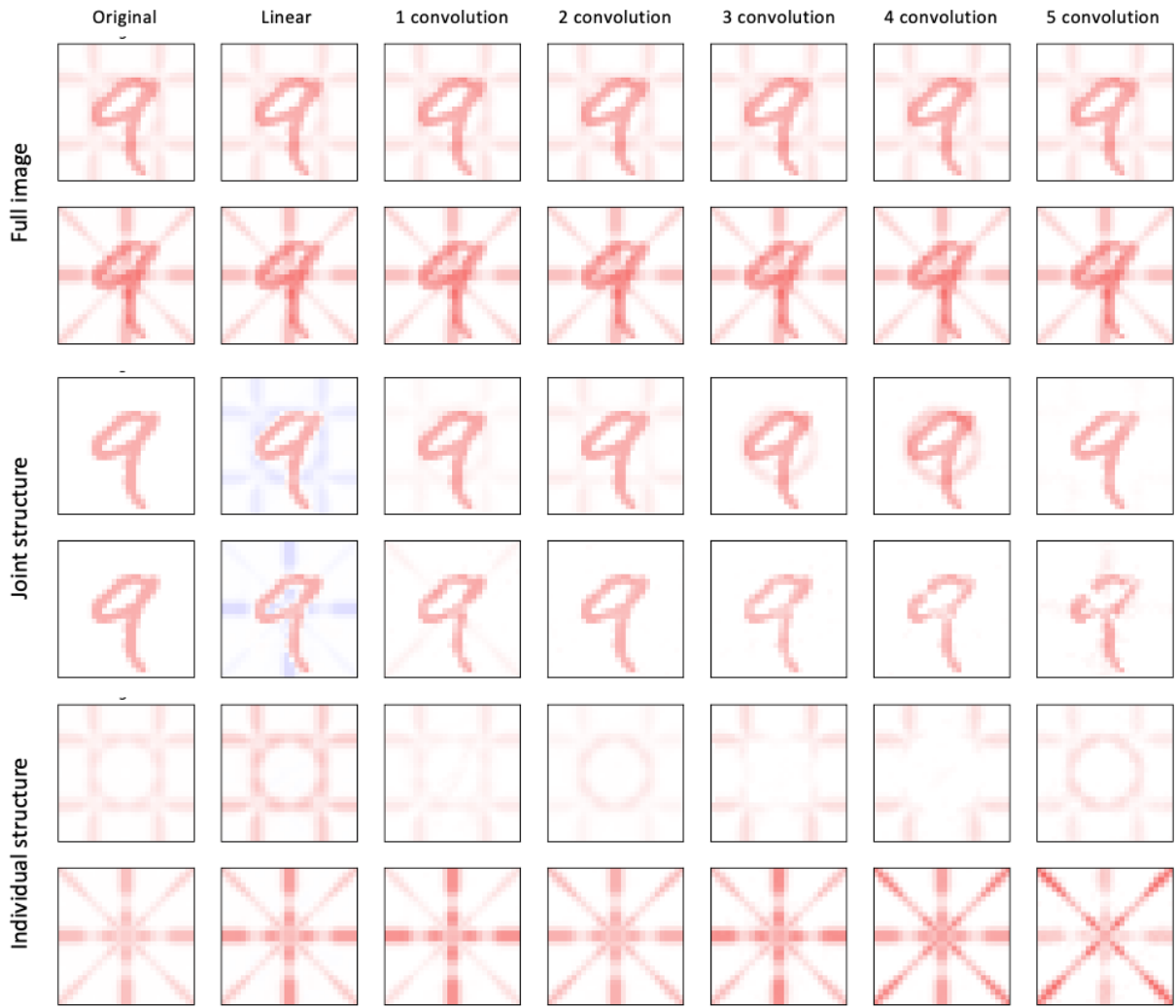

**Figure 7.** Visual representations of the input (top two rows, left column) to a merged DeepJIVE network and the constituent images representing its joint (middle rows, left column) and individual structures (bottom rows, left column), and the network output for explicit networks using either all latent space variables (top two rows), or just the joint or individual variables (middle and bottom rows, respectively). The networks that produced these images were trained without regression networks to enforce orthogonality.

### 3.2.2 MNIST paired dataset

To explore the potential advantages of non-linear and convolutional operations enabled by DeepJIVE on data analysis tasks, a merged network was trained with a regression network to reconstruct paired MNIST images so that the latent space variables could be used for a simple classification task. Because both images in each sample are of the same digit, the representation learned by the joint encoders should be related to the common aspects of each digit, and as the classification metrics of support vector machine classifiers trained on different sets of latent space variables in Figure 8 show, the ability for the DeepJIVE networks to capture these commonalities in the joint latent variables increases with the number of convolutional layers. While the accompanying images of the synthesized output show that this representation can be somewhat abstract, the improvement in accuracy, precision, recall, and F-1 score with the increase in layers indicates that non-linear, convolutional operations can cause more of the information related to identifying digits to be captured by the joint variables.

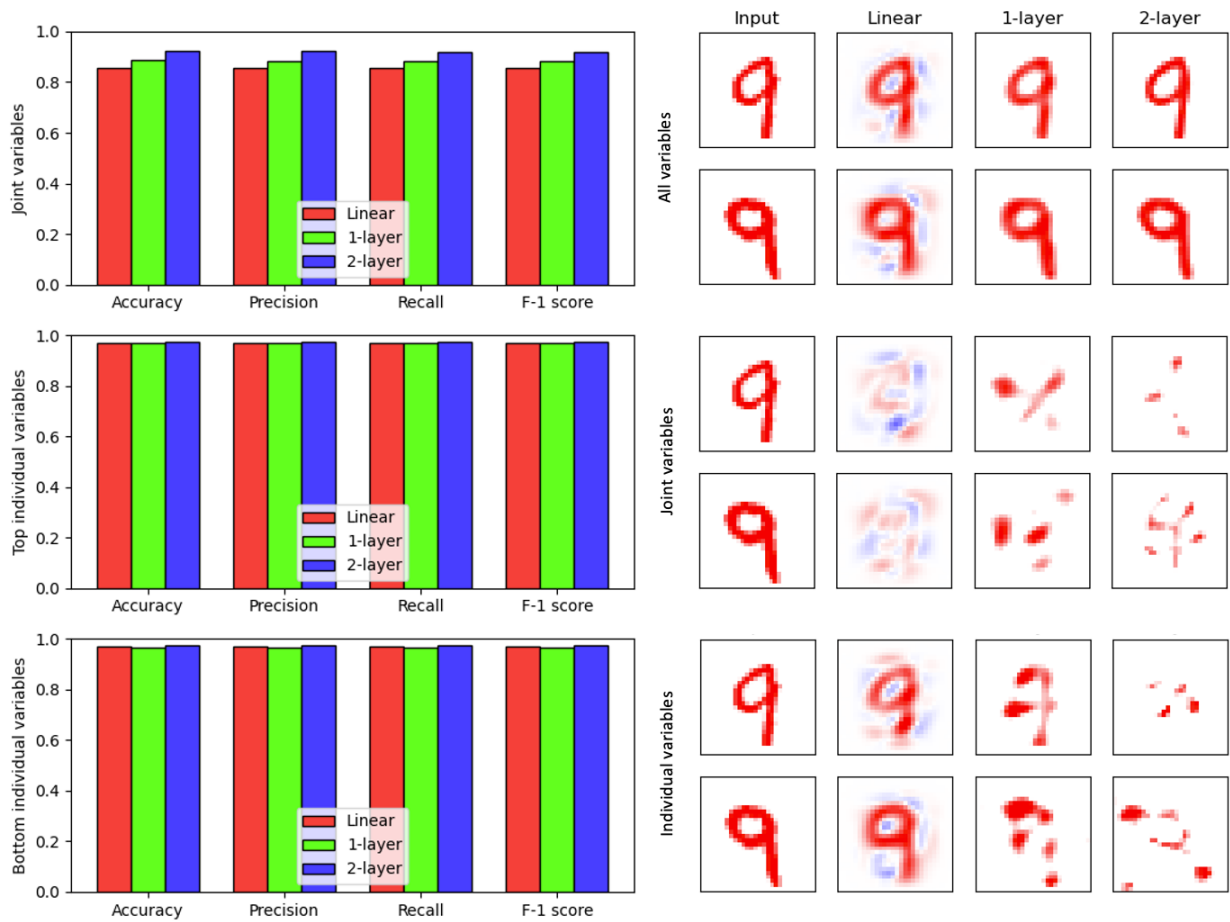

**Figure 8.** (Left) Performance metrics calculated for a SVM classifier trained on specified groups of variables from a merged network trained on the MNIST paired dataset. (Right) Images synthesized from specified groups of variables based on a given input (leftmost column) from the same networks that produced the latent space variables used for the classification task used to produce the bar graphs to the left.

### 3.3 Application of DeepJIVE to the ADNI dataset

To test the effectiveness of DeepJIVE on more complex data, a 3D convolutional explicit DeepJIVE network was trained on MRI and PET images from the ADNI dataset, and their usefulness in analyzing this dataset was assessed using a support vector machine (SVM) from the scikit-learn Python package[26] trained to identify CN and AD subjects from different groupings of the network's latent space variables. As the explicit network produces two sets of joint latent space variables, the joint variables derived from MRI and PET were each grouped with both sets of individual variables to form two separate datasets. These SVMs are compared in Figure 9 against an SVM classifier trained on the combined latent space variables from two non-DeepJIVE, 3D convolutional autoencoders using the same architecture and latent space variables as the constituent autoencoder modules of the DeepJIVE network. While all three classifiers have an accuracy, precision, recall, and F-1 score of at least 0.8, the classifiers trained on the DeepJIVE variables outperform the classifier trained on the variables from the separate autoencoders in all metrics.

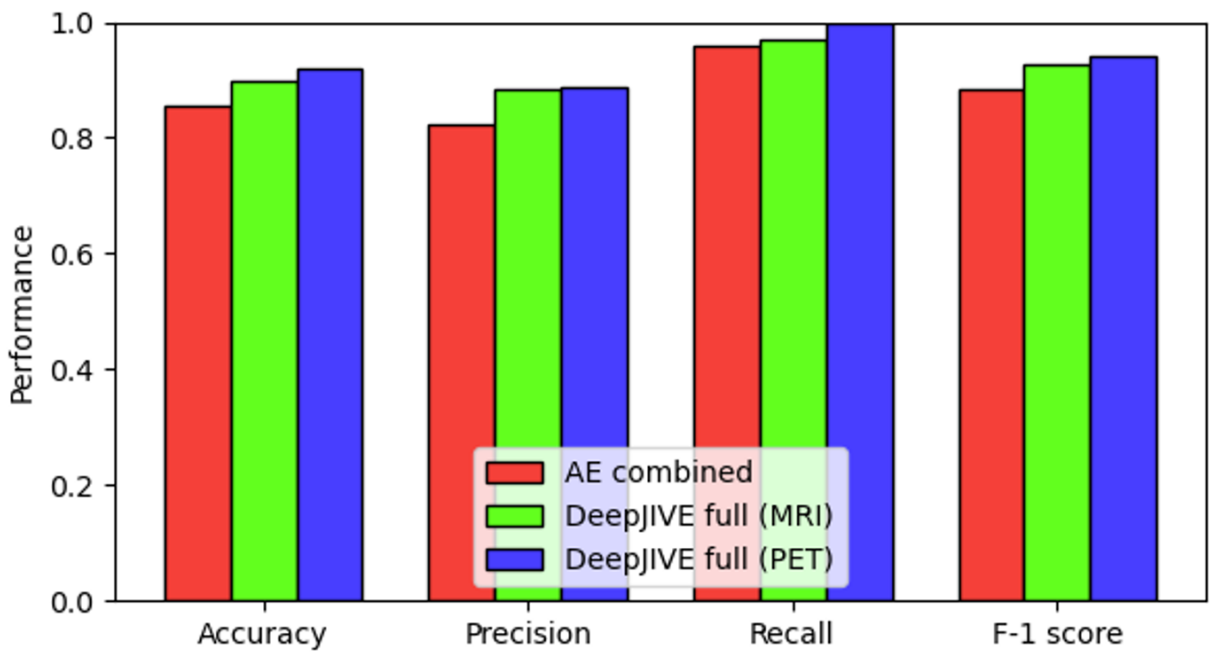

**Figure 9.** Classification accuracy, precision, recall, and F-1 score for SVM classifiers trained to classify cognitively normal subjects (CN) and subjects with dementia (AD) from the ADNI dataset using latent space variables from separate autoencoders (red) and an explicit DeepJIVE network (green and blue) trained on gray matter tissue probability maps (MRI) and AV45 PET images (PET).

More useful than performing mere classification is the potential to analyze the components of the network, especially the joint components, to understand how each data type covariates with each other. Figure 10 shows the T1w MRI image for a particular subject overlaid with a color map derived from a pseudo-saliency map for a specific component. The pseudo-saliency map was generated by encoding the subject's TPM and PET image, generating two images from the derived latent space after varying a single component by ±1 standard deviation of the whole dataset for that variable, and taking the difference between the two. Only voxels with a magnitude greater than the median of the absolute, non-zero intensity distribution of the difference image are colored. This component shown in Figure 10 was determined to have the greatest influence on classification performance from a permutation analysis, and as might be expected from the component identified as being the most relevant to identifying AD, it shows a negative correlation between TPM intensity on the external surfaces of the brain and intensity throughout the gray and white matter regions of the AV45 PET image. Since intensity in the TPM corresponds to tissue density, and intensity in the AV45 image is linked to Aβ accumulation, this component indicates that a correlation exists between cortical thinning and Aβ buildup in disease state, as is expected for those with AD. However, the component also shows a positive correlation between TPM intensity around the corpus callosum and intensity in the AV45 image, indicating that the more central regions of the brain degrade more slowly than the peripheral regions.

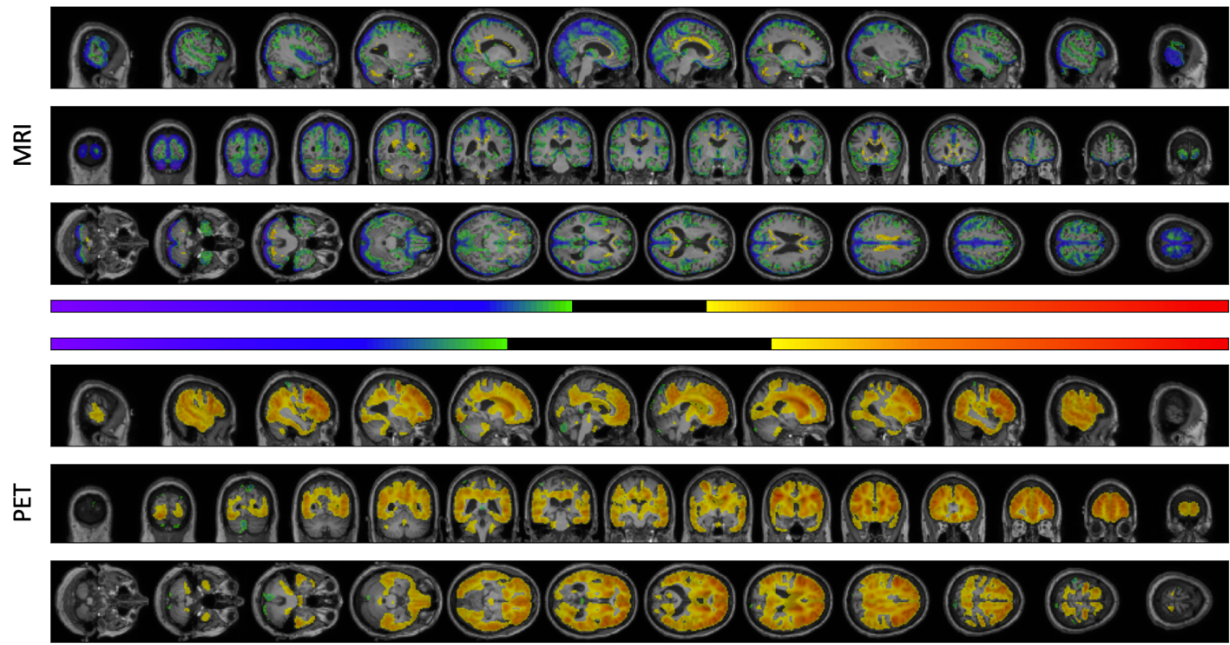

**Figure 10.** Difference image for a joint variable from the DeepJIVE network trained on gray matter TPM and AV45 images from the ADNI dataset overlaid on the T1w image from which the TPM was derived.

## 4. Discussion

We have proposed a deep-learning-based extension to JIVE that addresses the short-coming of conventional including being able to handle high-dimensional data such as images as input directly. Our extensive analyses on synthetic and real-world data have demonstrated that with careful considerations and design of strategies to achieve various requirements of JIVE, DeepJIVE can successfully uncover joint and individual variations of multimodal datasets. This allows for the study of covariations between multiple measures of the same biological system, such as between MR images and PET images as illustrated in real-world data application, between MR images and proteomics of plasma and/or cerebrospinal fluid, and others.

To faithfully achieve what JIVE does for low-dimensional data, DeepJIVE must fulfill several requirements: disentangle the joint and individual information from each other; produce joint latent variables that are equal to one another for each data type; and produce joint variables that are orthogonal to the individual variables. The ability for DeepJIVE to appropriately separate the joint and individual information can best be seen in the output produced by either the joint or individual variables shown in Figures 6 and 7 as the 1D and MNIST overlaid datasets were constructed to have a specific form for the joint and individual structures. The similarity between the DeepJIVE output and the constituent images used to construct each dataset, or image in the case of Figure 7, shows that the DeepJIVE networks are capable of identifying the elements of each dataset that covary with one another and will organize them into the correct set of variables. The ability to produce identical joint vectors from each data type can be seen in the visualizations of the weight vectors of the different DeepJIVE decoders in Figure 5A. Each type of network produced nearly identical joint structures for each dataset in Figure 6, and the left and right halves of each visualization are nearly identical, so the values of each variable used to produce those visualizations must be equal as well. While this is not perfect for more complex networks as demonstrated by the slight difference in performance between support vector classifiers trained using either the joint variables from the MRI joint encoder or the PET joint encoder in Figure 9, they are nonetheless sufficiently identical. The greatest hurdle DeepJIVE networks have to overcome with respect to faithfully implementing what JIVE does for low-dimensional data is the tendency of neural networks to find non-orthogonal solutions, but as can be seen in the reduced spread of the individual latent variables in Figure 5B for DeepJIVE networks trained with regression networks compared with those trained without, this hurdle can be overcome by providing additional gradients that steer the network towards orthogonality.

## 4.1 The role of identity constraint on separating joint and individual information

The natural tendency for DeepJIVE networks to separate joint and individual information is a direct result of its ability to produce identical joint variables from each encoder, and why autoencoders tend to find non-orthogonal latent spaces. A linear decoder can be thought of as a $p_k$-dimensional hyperplane composed of r vectors, $\vec{w}_\tau$, that minimizes the average distance between the initial data, $X$, and its projection onto the hyperplane, $\hat{X}$, when parameterized by a corresponding set of variables, $\lambda_\tau$, learned by the encoder. When describing hyperplanes using a set of vectors, it is customary to use unit vectors for $\vec{w}_\tau$ and let $\hat{X}$ be defined by the distribution of $\lambda_\tau$, but any set of non-parallel vectors that lie within a hyperplane, regardless of direction or magnitude, can be used to define that hyperplane, and the non-perpendicularities of the defining vectors can be compensated by altering the distributions of the aligned vectors' parameters. When the decoder is initialized, the vectors comprising the decoder are randomly distributed in terms of magnitude and direction, and training will cause them to change magnitude and direction until they can adequately produce $\hat{X}$ from the distributions of $\lambda_\tau$ learned by the encoder. However, once all $\vec{w}_\tau$ lie within the optimal hyperplane and the distributions of their corresponding $\lambda_\tau$ match the directions and magnitudes of $\vec{w}_\tau$ to adequately produce $\hat{X}$, there is no incentive in terms of the loss function to find a more orthogonal representation. However, if the magnitude and direction of a single vector and the mean and spread of its parameter's distribution were to be fixed to some value, then the remaining $\vec{w}_\tau$ would need to change in magnitude and direction to compensate. While not fixed, the requirement that the values of the joint latent spaces be equal to each other effectively provides an external pressure on the distributions of $\lambda_J$, causing their $\vec{w}_{J_k}$ to rotate to lie within the optimal hyperplane defined by factors common to all datatypes, and their $\vec{w}_{S_k}$ rotate to compensate. However, this intrinsic pressure is still not of orthogonality, but general direction; it is clearly possible for there to be some degree of dependence between the variables of each structure as seen in Figures 5 and 7. In order to achieve true orthogonality, the distributions of $\lambda_J$ and $\lambda_{S_k}$ must be set to an ideal distribution that they would have if all components were orthogonal, and this is the purpose of the proposed regression network. It determines the deviation of the $\lambda_{S_k}$ distribution caused by any alignment between the constituent sub-functions of $f_{J_k}^D(\lambda_J)$ and $f_{S_k}^D(\lambda_{S_k})$ so that it can be removed.

### 4.2 Interpreting DeepJIVE results

While the improvement in classification metrics seen in Figure 8 show that convolutional layers can improve the power of the learned joint latent variables to perform certain downstream tasks, it is important to note that the joint latent variables do not necessarily contain all information that is necessary for the task, or even all information that one might assume should be considered to be governed by joint factors. The joint variables only govern variation that is common between the different datatypes, but as illustrated in the joint and individual reconstructions of the linear network in Figure 8, this co-variation may only exist in the lower-order components, meaning that the individual components may be much more descriptive of the dataset. Additionally, improvements in the power of the joint variables do not necessarily come at the expense of the power of the individual variables, so one should not disregard the information contained in the individual structure. The greater performance of the entire DeepJIVE variable set over the combined variable sets from normally trained, separate autoencoders in Figure 9 indicates that the disentanglement of mutual information from individual information can provide some improvement in the descriptive power of the variables found by deep learning, but this improvement applies across all variables, not just the joint or individual. One should also be mindful of interpreting the components by visualization. As seen in the reconstructions of the MNIST digits using only joint or individual variables in Figures 8 and the larger networks of Figure 7, excluding an entire set of variables from a non-linear function can produce results that are non-sensical to the human eye even though they are produced from variables that are objectively useful in other tasks. The power of the DeepJIVE network is to be able to produce images from the mutual information between different data types contained in the joint structure to see how each datatype changes in response to the other, but this should be done using difference images similar to the one shown in Figure 10.

### 4.3 Comparison to similar work

These architectures are similar to the FR-DCCA architecture proposed by Sun et al. to learn CCA, but the primary difference lies in the ways in which correlation is enforced between the joint encoders.[27] Specifically, FR-DCCA seeks to maximize correlation between the output of the shared encoders by essentially performing CCA on the output,[28] which requires computing the gradient across the entire dataset or sufficiently large minibatches before updating the networks through backpropagation. DeepJIVE seeks to not merely maximize correlation but enforce identity between its equivalent of the shared information encoders, allowing for smaller batch sizes. FR-DCCA also attempts to minimize the information shared between the shared and specific encoders by adding the Jensen-Shannon divergence between their outputs to the loss function while DeepJIVE uses an additional adversarial network in a GAN-like scheme to estimate and remove the nominal variance caused by shared variation.

## 5. Conclusion

This work presents DeepJIVE, a deep learning extension of JIVE to allow direct handling of high-dimensional datasets and uncover non-linear covariation patterns. It uses a set of parallel autoencoders to separate the variation within multiple data types caused by factors common and unique to each data type. This potentially improves upon traditional JIVE implementations by using non-linear and convolutional layers to transform non-linear relationships into linear relationships such that JIVE can be performed. Three different types of DeepJIVE networks that are differentiated based on the number of autoencoders used for the joint structure and training loss function were tested, and all three showed an adequate ability to learn JIVE based on their performance on synthetic and real-world datasets. As demonstrated by the MNIST dataset, their performance was affected by the number of convolutional layers in addition to traditional JIVE parameters such as the rank of the decomposition.

**Acknowledgements**
This work is supported by NIH (awards: R01AG089806, R01AG071174, R01AG070937). Data collection and sharing for this project was funded by the Alzheimer's Disease Neuroimaging Initiative (ADNI) (National Institutes of Health Grant

U01 AG024904) and DOD ADNI (Department of Defense award number W81XWH-12-2-0012). ADNI is funded by the National Institute on Aging, the National Institute of Biomedical Imaging and Bioengineering, and through generous contributions from the following: AbbVie, Alzheimer's Association; Alzheimer's Drug Discovery Foundation; Araclon Biotech; BioClinica, Inc.; Biogen; Bristol-Myers Squibb Company; CereSpir, Inc.; Cogstate; Eisai Inc.; Elan Pharmaceuticals, Inc.; Eli Lilly and Company; EuroImmun; F. Hoffmann-La Roche Ltd and its affiliated company Genentech, Inc.; Fujirebio; GE Healthcare; IXICO Ltd.;Janssen Alzheimer Immunotherapy Research & Development, LLC.; Johnson & Johnson Pharmaceutical Research & Development LLC.; Lumosity; Lundbeck; Merck & Co., Inc.;Meso Scale Diagnostics, LLC.; NeuroRx Research; Neurotrack Technologies; Novartis Pharmaceuticals Corporation; Pfizer Inc.; Piramal Imaging; Servier; Takeda Pharmaceutical Company; and Transition Therapeutics. The Canadian Institutes of Health Research is providing funds to support ADNI clinical sites in Canada. Private sector contributions are facilitated by the Foundation for the National Institutes of Health (www.fnih.org). The grantee organization is the Northern California Institute for Research and Education, and the study is coordinated by the Alzheimer's Therapeutic Research Institute at the University of Southern California. ADNI data are disseminated by the Laboratory for Neuro Imaging at the University of Southern California.